%% file: PaperForReview.tex
\documentclass[10pt,twocolumn,letterpaper]{article}

\usepackage{cvpr}              

\usepackage{graphicx}
\usepackage{amsmath}
\usepackage{amssymb}
\usepackage{booktabs}
\usepackage{multirow}
\usepackage{multicol}
\usepackage{enumitem}
\usepackage[accsupp]{axessibility}

\usepackage[pagebackref,breaklinks,colorlinks]{hyperref}

\usepackage[capitalize]{cleveref}
\crefname{section}{Sec.}{Secs.}
\Crefname{section}{Section}{Sections}
\Crefname{table}{Table}{Tables}
\crefname{table}{Tab.}{Tabs.}

\def\cvprPaperID{4} 
\def\confName{CVPR}
\def\confYear{2023}

\begin{document}

\title{Are we certain it's anomalous?}

\author{Alessandro Flaborea$^{1}$\space\space\space Bardh Prenkaj$^1$ \space\space\space Bharti Munjal$^2$ \space\space\space Marco Aurelio Sterpa$^1$ \space\space\space \\ Dario Aragona$^1$ \space\space\space Luca Podo$^1$ \space\space\space Fabio Galasso$^1$\\
$^1$ Sapienza University of Rome, Italy \space\space\space $^2$ Technical University of Munich, Germany\\
{\tt\small \{flaborea,prenkaj,aragona,podo,galasso\}@di.uniroma1.it, munjalbharti@gmail.com}}
\maketitle

\input{sec/abstract}

\input{sec/intro}

\input{sec/related}

\input{sec/method}

\input{sec/experiments}

\input{sec/conclusion}

{\small
\bibliographystyle{ieee_fullname}
\bibliography{references}
}

\end{document}

%% file: sec/abstract.tex
\begin{abstract}
The progress in modelling time series and, more generally, sequences of structured data has recently revamped research in anomaly detection. The task stands for identifying abnormal behaviors in financial series, IT systems, aerospace measurements, and the medical domain, where anomaly detection may aid in isolating cases of depression and attend the elderly. Anomaly detection in time series is a complex task since anomalies are rare due to highly non-linear temporal correlations and since the definition of anomalous is sometimes subjective.

Here we propose the novel use of Hyperbolic uncertainty for Anomaly Detection (HypAD). HypAD learns self-supervisedly to reconstruct the input signal.
We adopt best practices from the state-of-the-art to encode the sequence by an LSTM, jointly learned with a decoder to reconstruct the signal, with the aid of GAN critics. Uncertainty is estimated end-to-end by means of a hyperbolic neural network. By using uncertainty, HypAD may assess whether it is certain about the input signal but it fails to reconstruct it because this is anomalous; or whether the reconstruction error does not necessarily imply anomaly, as the model is uncertain, e.g. a complex but regular input signal. The novel key idea is that a \emph{detectable anomaly} is one where the model is certain but it predicts wrongly.
HypAD outperforms the current state-of-the-art for univariate anomaly detection on established benchmarks based on data from NASA, Yahoo, Numenta, Amazon, and Twitter. It also yields state-of-the-art performance on a multivariate dataset of anomaly activities in elderly home residences, and it outperforms the baseline on SWaT. 
Overall, HypAD yields the lowest false alarms at the best performance rate, thanks to successfully identifying detectable anomalies.
\vspace*{-3mm}

\end{abstract}

%% file: sec/intro.tex
\section{Introduction}
\label{sec:intro}

\input{fig/teaser}
Anomaly detection stands for detecting outliers (anomalies) in data, i.e.\ points that deviate significantly from the distribution of the data. Outlier detection, however, is an under-specified and consequently ill-posed task due to its inherent unsupervised nature. Anomaly detection strategies such as distance-based \cite{fan2006nonparametric,ghoting2008fast}, density-based \cite{lof,papadimitriou2003loci}, and subspace-based methods \cite{keller2012hics,lazarevic2005feature} have been pioneers in the literature. Additionally, autoencoders \cite{chen2017outlier,sarvari2021unsupervised}, and adversarial networks \cite{geiger2020tadgan} have given a substantial contribution. However, the literature neglects assessing the trustfulness of the predicted outcomes, namely their uncertainty.

Uncertainty is a measure of model confidence, which may be learnt from the data~\cite{lakshminarayanan2016simple,kendall2017uncertainties} or by the use of extra instances~\cite{gal2016dropout}. Uncertainty estimation has been a long-standing challenge in machine learning. Most recently, it has been successfully adopted to improve performance on object detection~\cite{jiang2018acquisition,kuppers2020multivariate,9156274,neumann2018relaxed}, pose estimation~\cite{vakhitov2021uncertainty}, unsupervised and self-supervised learning~\cite{fiery2021,8289350,suris2021hyperfuture}. Yet, end-to-end and data-driven uncertainty estimation for anomaly detection remains a complex task. This complexity is further exacerbated by the rarity of anomalous events which, generally, are not present during the learning (training) phase - i.e., open set problem - leading deep learning models to be overconfident when predicting them \cite{Hein_2019_CVPR}.

In this work, we propose a novel model based on Hyperbolic uncertainty for Anomaly Detection, which we dub HypAD. We leverage the current state-of-the-art technique for anomaly detection in univariate time series, TadGAN~\cite{geiger2020tadgan}. TadGAN detects anomalies by attempting to reconstruct the input signal, making use of an LSTM sequence encoding and two GAN critics, cf.\ Sec.~\ref{sec:background}. We introduce uncertainty into the anomaly detector: i.e., we map the input and the reconstructed signal into a hyperbolic space, where the signals additionally have an uncertainty score; and we train the novel embeddings end-to-end with a Poincaré distance loss, cf.\ Sec.~\ref{sec:hypAD}.

HypAD uses uncertainty to discern whether the reconstruction error is large because the signal is anomalous, or simply because the model cannot reconstruct it well. In the former case, HypAD is certain about the reconstruction (e.g.\ most signal is well-behaved and the model expects known patterns) but its reconstruction is wrong, as a part of the signal is anomalous. In the latter, HypAD downgrades its anomaly score because it is not certain about signal reconstruction. This may be because of a complex pattern, which the model did not have enough capacity to learn. The larger uncertainty indicates that the larger reconstruction error may be due to an anomaly or to a model failure in the reconstruction. (See discussion in Sec.~\ref{sec:hypUn}.)

Thanks to uncertainty, HypAD outperforms the state-of-the-art univariate anomaly detector TadGAN~\cite{geiger2020tadgan} on the established univariate benchmarks of NASA, Yahoo, Numenta Anomaly Benchmark~\cite{lavin2015evaluating}, as well as on two multivariate datasets of daily activities in elderly home residences CASAS~\cite{cook2012casas} and industrial water treatment plant SWaT~\cite{Mathur2016SWaTAW}. As we show in experimental results in Sec.~\ref{sec:expres}, reducing anomaly scores in uncertain cases also yields fewer false alarms (the model achieves the best F1 performance with larger precision).

The main contributions of this work are:
\begin{itemize}[noitemsep,topsep=0pt]
\item We propose the first model for anomaly detection based on hyperbolic uncertainty;
\item We propose the novel key idea of \emph{detectable anomaly}: an instance is anomalous when the model is certain about it but wrong;
\item We integrate the estimated uncertainty into a state-of-the-art univariate anomaly detector and outperform it on established univariate and multivariate datasets.
\end{itemize}

%% file: fig/teaser.tex
\begin{figure}[t] 
\vspace*{-2mm}
\begin{center}
	\includegraphics[trim=0cm 0cm 0cm 0cm, clip=true, width=.78\linewidth]{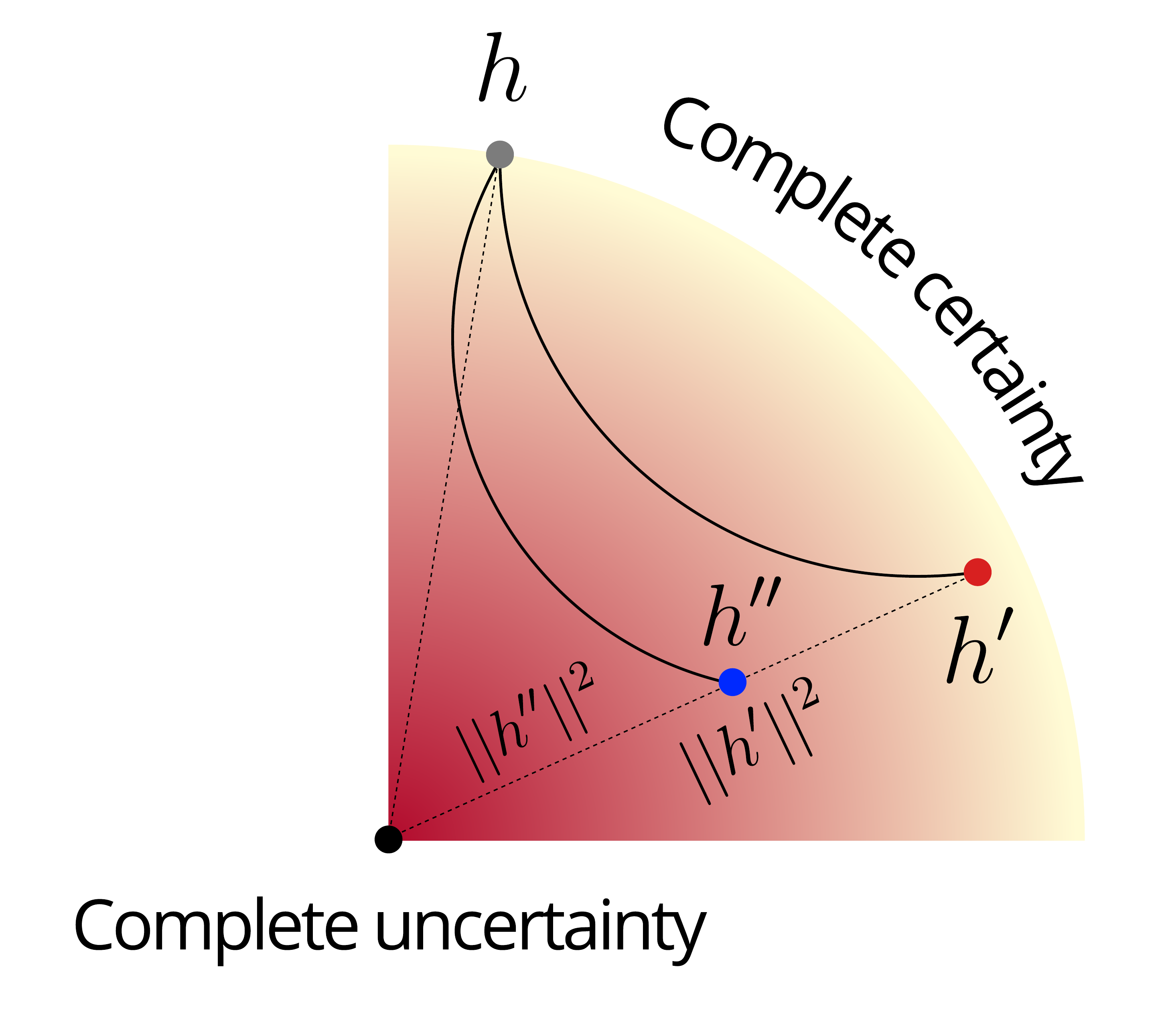}
\end{center}
\vspace*{-7mm}
\caption{HypAD detects anomalies by the joint use of reconstruction error and uncertainty, learning both aspects end-to-end. In the illustration, the colored circular sector represents the hyperbolic Poincaré ball, where the radius distance of data embeddings is their degree of certainty (points on the circumference are most certain).
$\boldsymbol{h}$ is the hyperbolic mapping of the input signal, which HypAD attempts to match by the reconstruction $\boldsymbol{h'}$. Thanks to hyperbolic neural networks and their exponentially larger penalization for errors at high certainty, HypAD learns to prefer signal reconstructions such as $\boldsymbol{h''}$, i.e.\ with the same amount of error as $\boldsymbol{h'}$ (the same angle and cosine distance) but smaller radius $\|\boldsymbol{h''}\|^2$ and thus higher uncertainty. HypAD uses both the reconstruction error and the uncertainty to identify detectable anomalies, where the model is certain but the prediction is wrong, i.e.\ it knows what to expect but something anomalous occurs.
}
\vspace*{-4mm}
%
%
%
\label{fig:teaser}
\end{figure}

%% file: sec/related.tex
\section{Related works}
\label{sec:related}

To the best of our knowledge, this is the first work to have combined anomaly detection with end-to-end and data-driven uncertainty estimation and the first work to have further proposed hyperbolic uncertainty for it. Previous work relates to ours from three main perspectives, which we review here: uncertainty estimation techniques, anomaly detection in time series, and hyperbolic neural networks.

\subsection{Uncertainty Estimation Techniques}\label{sec:related-uncertainty}

We identified two different strategies for approximating uncertainty. \textit{Ensemble-based posterior approximation} uses several weak models to make naive predictions and combine them according to a consensus function into a more complex predictive model \cite{dietterich2000ensemble}. One of the most popular approaches to uncertainty estimation based on ensembles is Monte Carlo (MC) Dropout. It drops neurons on every layer during training and test phases \cite{gal2016dropout}. 

\textit{Generative models for aleatory modelling} use an additional latent variable to make stochastic predictions and evaluate the uncertainty of the model. Generative Adversarial Networks (GANs) \cite{goodfellow2014generative} play a min-max game where the discriminator needs to distinguish between real examples and the generated outcomes. GANs have state-of-the-art performances and we build on top of that by attaching hyperspace mapping layers to estimate the uncertainty of the model. Another interesting approach to estimate uncertainty is by using energy-based models \cite{du2019implicit,salakhutdinov2009deep,xie2016theory}. They learn an energy function that models the compatibility of the input and the output. Our method transcends energy-based models because the integrated hyperbolic uncertainty mechanism does not suffer from cold- nor warm-start problems \cite{zhang2021dense} which undermine the training complexity \cite{xie2018cooperative}.

Besides the above methods of approximating uncertainty, in the past, baselines like Hotelling's T$^2$ \cite{hotelling1992generalization} exploited the Mahalanobis distance that incorporates uncertainty by measuring the (co-)variance, with the deviation (i.e., being anomalous or not). Nevertheless, current deep learning models in anomaly detection do not exploit uncertainty to calibrate their overconfident predictions for unseen data (anomalies) that lie outside the manifold of normal instances. Hence, an end-to-end and data-driven uncertainty estimation technique for anomaly detection is necessary.

\subsection{Anomaly Detection in Time Series}\label{sec:related_anomaly}
We identified five categories of methods proposed in the literature for anomaly detection in time series. \textit{Distance-based} outlier detectors consider the distance of a point from its k-nearest neighbours \cite{conf/vldb/KnorrN98, angiulli2002fast,ghoting2008fast}. \textit{Density-based} methods~\cite{lof,papadimitriou2003loci,He2003DiscoveringCL, 10.1145/3447548.3467137, 10.1145/3292500.3330672,WangCNLCT21, su2019robust} take into account the density of the point and its neighbours. \textit{Prediction-based methods} \cite{benkabou2021local,unsupahmad,conf/kdd/HundmanCLCS18} calculate the difference between the predicted value and the true value to detect anomalies. \textit{Reconstruction based} methods \cite{An2015VariationalAB,Malhotra2016LSTMbasedEF,10.1145/3447548.3467174} compare the input signal and the reconstructed one in the output layer typically by using autoencoders. These methods assume that anomalies are difficult to reconstruct and are lost when the signal gets mapped to lower dimensions. Thus, a higher reconstruction error means a higher anomaly score. \cite{Malhotra2016LSTMbasedEF} uses an LSTM autoencoder for multi-sensor anomaly detection. \cite{chen2017outlier,sarvari2021unsupervised} use an ensemble of autoencoders to boost performances by focusing on learning the inlier characteristics at each iteration.~\cite{li2021multivariate} uses a hierarchical variational auto-encoder with two stochastic latent variables to learn the temporal and inter-metric embeddings for multivariate data.SISVAE~\cite{li2020anomaly} uses a variational auto-encoder with a smoothness-inducing prior over possible estimations to capture latent temporal structures of time series without relying on the assumption of constant noise. Recently, \textit{GANs} have been employed to detect anomalies in time series data. Our method also lies in this category. MAD-GAN \cite{Li2019MADGANMA} combines the discriminator output and reconstruction error to detect anomalies in multivariate time series. BeatGAN~\cite{ijcai2019-616} uses an encoder-decoder generator with a modified time-warping-based data augmentation to detect anomalies in medical ECG inputs. TadGAN~\cite{geiger2020tadgan} uses a cycle-consistent GAN architecture with an encoder-decoder generator and proposes several ways to compute reconstruction error and its combination with the critic outputs. We build on top of TadGAN's architecture by incorporating the hyperbolic mapping layer to the reconstructed time-windows to assess the uncertainty of the detector. 




\subsection{Hyperbolic Neural Networks}\label{sec:related_hnn}
Deep representation learning in hyperspaces has gained momentum after the pioneering work of hyperNNs~\cite{NEURIPS2018_dbab2adc} that generalizes Euclidean operations 
to their counterparts in hyperspace. The authors propose analog counterparts in the hyperspace of neural network components such as fully connected (FC) layers, multinomial logistic regression (MLR), and recurrent neural networks. Furthermore, methods like Einstain midpoint~\cite{gulcehre2018hyperbolic} and Fréchet mean~\cite{Lou2020DifferentiatingTT} propose different ways of aggregating features in hyperspace. The work in \cite{shimizu2021hyperbolic} extends hyperNN and proposes Poincaré split/concatenation operations, generalizing the convolutional layer to hyperspace.~\cite{NEURIPS2019_103303dd,chami2019hyperbolic} propose hyperbolic graph neural networks, leveraging hyperNNs. 

Thus formulated, hyperNNs have mainly been adopted to improve performance by leveraging hierarchies and uncertainty in zero-shot learning \cite{Liu_2020_CVPR}, re-identification \cite{Khrulkov_2020_CVPR}, and action recognition \cite{9157196}. Of particular interest, \cite{suris2021hyperfuture} has leveraged hyperNNs to model a hierarchy of actions from unlabeled videos. To the best of our knowledge, this is the first work to have applied hyperNNs for sequence modelling with the goal of anomaly detection.

%% file: sec/method.tex
\section{Method}
In this section, we first discuss best practises in anomaly detection (Sec.~\ref{sec:background}); then we detail the proposed hyperbolic uncertainty and its use for detecting anomalies (Sec.~\ref{sec:hypAD}); finally, we discuss the motivation for it (Sec.~\ref{sec:mot}).

\input{fig/hyperTadGAN.tex}

\subsection{Background}
\label{sec:background}
\input{fig/cosine_barplot.tex}

The current state-of-the-art in univariate anomaly detection is a reconstruction-based technique~\cite{geiger2020tadgan} which additionally leverages a GAN critic score. TadGAN encodes the input data $x$ to a latent space and then decodes the encoded data. This encoding-decoding operation requires two mapping functions $E : X \rightarrow Z$ and $G : Z \rightarrow  X$.
The reconstruction operation can be given as $x \rightarrow E(x) \rightarrow G(E(x)) \rightarrow \widetilde{x} \approx x$. TadGAN leverages adversarial learning to train the two mappings by using two adversarial critics $D_x$ and $D_z$. The goal of $D_x$ is to distinguish between the real and the generated time series, while $D_z$ measures the performance of the mapping into latent space. The model is trained using the combination of Wasserstein loss~\cite{pmlr-v70-arjovsky17a} and Cycle consistency loss~\cite{8237506}. 
TadGAN computes the reconstruction error $RE(x)$ between x and $\widetilde{x}$ using
three types of reconstruction  functions: \textbf{i.}\ Point-wise difference: considers the difference of values at every timestamp; \textbf{ii.}\ Area difference: is applied on signals of fixed lengths and measures the similarity between local regions; \textbf{iii.}\ Dynamic time warping: additionally handles time gaps between the two signals to calculate the reconstruction error. To calculate the anomaly score, TadGAN first normalises the reconstruction error and critic scores by subtracting the mean and dividing by standard deviation. The normalised scores, $Z_{RE}(x)$ and $Z_{D_x}(x)$, are then combined using their product:
\begin{equation}
    s_p(x) = Z_{RE}(x) \cdot Z_{D_x}(x)
    \label{eq:anomaly_score_prod}
\end{equation}



\subsection{Hyperbolic uncertainty for Anomaly detection (HypAD)}
\label{sec:hypAD}

We propose a novel model for anomaly detection in time series based on hyperbolic uncertainty. HypAD is a reconstruction-based model and minimises the reconstruction loss, given by a measure of the hyperbolic distance between the input signal and its reconstruction. In hyperbolic space, errors are exponentially larger when predictions are certain. Therefore, HypAD tends to predict either certain correct reconstructions or uncertain possibly mistaken reconstructions. This leads, as we discuss in Sec.~\ref{sec:mot}, to a novel definition of detectable anomaly: i.e. the case of a large reconstruction error with high certainty.

\subsubsection{Hyperbolic Reconstruction Error}
\label{sec:hypRe}

The proposed HypAD, illustrated in Fig.~\ref{fig:hypAD}, integrates hyperbolic neural networks into the reconstruction-based architecture of TadGAN. In HypAD the input signal $x$ is first passed through an encoder, then followed by a decoder sub-network. The output of the decoder $G(E(x))$ as well as the original signal $x$ are mapped to the hyperspace, shown as the red edge box with red background. 

An $n$-dimensional hyperspace is a Riemannian geometry with a constant negative sectional curvature. As in~\cite{Khrulkov_2020_CVPR,suris2021hyperfuture}, we adopt the Poincaré ball model of hyperspaces, given by the manifold $\mathbb{D}_n = \{x \in R^n : \|x\| < 1\}$ endowed with the Riemannian metric $g^\mathbb{D}(x) = \lambda_x^2g^E$,
where $\lambda_x =\frac{2}{1-\|x\|^2}$ is the \textit{conformal factor}
and $g^E = I_n$ is the Euclidean
metric tensor. For details, see~\cite{riemannian,smoothmanifold}.
\input{tab/datasets}

To map $x$ and $G(E(x))$ to the Poincaré ball, we leverage an \textit{exponential map} centered at $\bf{0}$~\cite{suris2021hyperfuture}. This is followed by a hyperbolic feed-forward layer~\cite{NEURIPS2018_dbab2adc} to estimate the corresponding hyperbolic embeddings $h$ and $\widetilde{h}$, solid green boxes in Fig.~\ref{fig:hypAD}. Finally, the two hyperbolic embeddings are compared using the Poincaré distance, formulated as follows:
\begin{equation}
{RE}(x)  =  \cosh^{-1}\Bigg(1 + 2{\frac{\|h-\widetilde{h}\|^2}{\big(1-\|h\|^2\big)\big(1-\|\widetilde{h}\|^2\big)}\Bigg)}
\label{eq:poincare}
\end{equation}
where $\|h\|^2$ and $\|\widetilde{h}\|^2$ are the distances of the embeddings from the center of the Poincaré ball. The same reconstruction error function $RE(x)$, is used at train and inference.

\subsubsection{Hyperbolic Uncertainty } 
\label{sec:hypUn}
A key property of the Poincaré ball is that the distance between two points grows exponentially as we move away from the origin. This means that an erroneous reconstruction towards the circumference of the disk is penalised exponentially more than an erroneous reconstruction close to the centre. This leads to the useful tendency of HypAD to either predict a matched reconstruction ($\widetilde{h}$ and $h$ are close by) or an unmatched reconstruction towards the origin ($\widetilde{h}$ and $h$ are far away, $\|h\|^2$ and $\|\widetilde{h}\|^2$ are small), in order to minimize Eq. \eqref{eq:poincare}. Hence, the distance of the reconstruction to the origin provides a natural estimate of the model's uncertainty, referred to as hyperbolic uncertainty, $U(x)$, thus formulated:
\begin{equation}
{U}(x)  =  1 - \|   \tilde{h}\|^2
\label{eq:hypUn}
\end{equation}
The smaller the distance from the origin, the more uncertain the model.

\subsubsection{Combining Hyperbolic Uncertainty with Reconstruction Error and Critic Score}
\label{sec:combine}

Hyperbolic uncertainty ${U}(x)$ is integrated into the anomaly score as follows:
\begin{equation}
    s_u(x) =  Z_{RE}(x) \cdot Z_{D_x}(x) \cdot (1-{U}(x))
    \label{eq:uncer}
\end{equation}
Eq.~\eqref{eq:uncer} brings together the reconstruction error $Z_{RE}(x)$ (the larger the error, the more likely the anomaly) with the critic score $Z_{D_x}(x)$ (larger critic scores point to anomalies) and the model certainty: $1-{U}(x)$. The simple multiplication formulation of the certainty of the model reduces the scores of anomalies when HypAD is not confident of the reconstructions. While being simple, this outperforms the current state-of-the-art, as we show in Sec.~\ref{sec:expres}.

\subsection{Motivation for HypAD}
\label{sec:mot}

HypAD takes motivation from a key idea: \textit{a detectable anomaly is one where the model is certain, but it predicts wrongly}. In other words, if the model encounters a known pattern, which it knows how to reconstruct, then it will call it anomalous if the reconstruction does not match the input signal. The principled formulation of hyperbolic uncertainty is critical towards this goal: HypAD predicts a reconstruction as uncertain if it is doubtful that it may be wrong.

Fig.~\ref{fig:cosine_barplot} illustrates this key concept for all the datasets. The first, second and third rows correspond to the univariate, U-CASAS and SWaT datasets respectively.
The bar plot depicts the average cosine distances between the input signals and their reconstructions, against specific intervals of uncertainty, along the x-axis.
The higher the cosine distances, the more distinct the reconstruction is from the provided signals\footnote{The Poincaré ball model is conformal to the Euclidean space and it preserves the same angles~\cite{shimizu2021hyperbolic}.}. 
Note that for the first two rows, the initial three plots (columns) correspond to the signals that report the best improvement in F1-score and the last plot corresponds to the signal with the worst improvement. A single bar plot is reported for SWaT because this consists of a single long-term signal. Observe, in the second row, how HypAD learns to correctly assign higher uncertainty to wronger estimates for the cases of Fall, Weakness, and Nocturia. For the last signal SlowerWalking, HypAD fails to learn a meaningful uncertainty and labels all reconstructions as certain. It is however notable that the representation is still interpretable and the case of failure discernible. Trends are similar for the cases of Univariate and SWaT datasets. 




%


%% file: fig/hyperTadGAN.tex
\begin{figure}[t] 
\begin{center}
	\includegraphics[trim=0cm 0cm 0cm 0cm, clip=true, width=\linewidth]{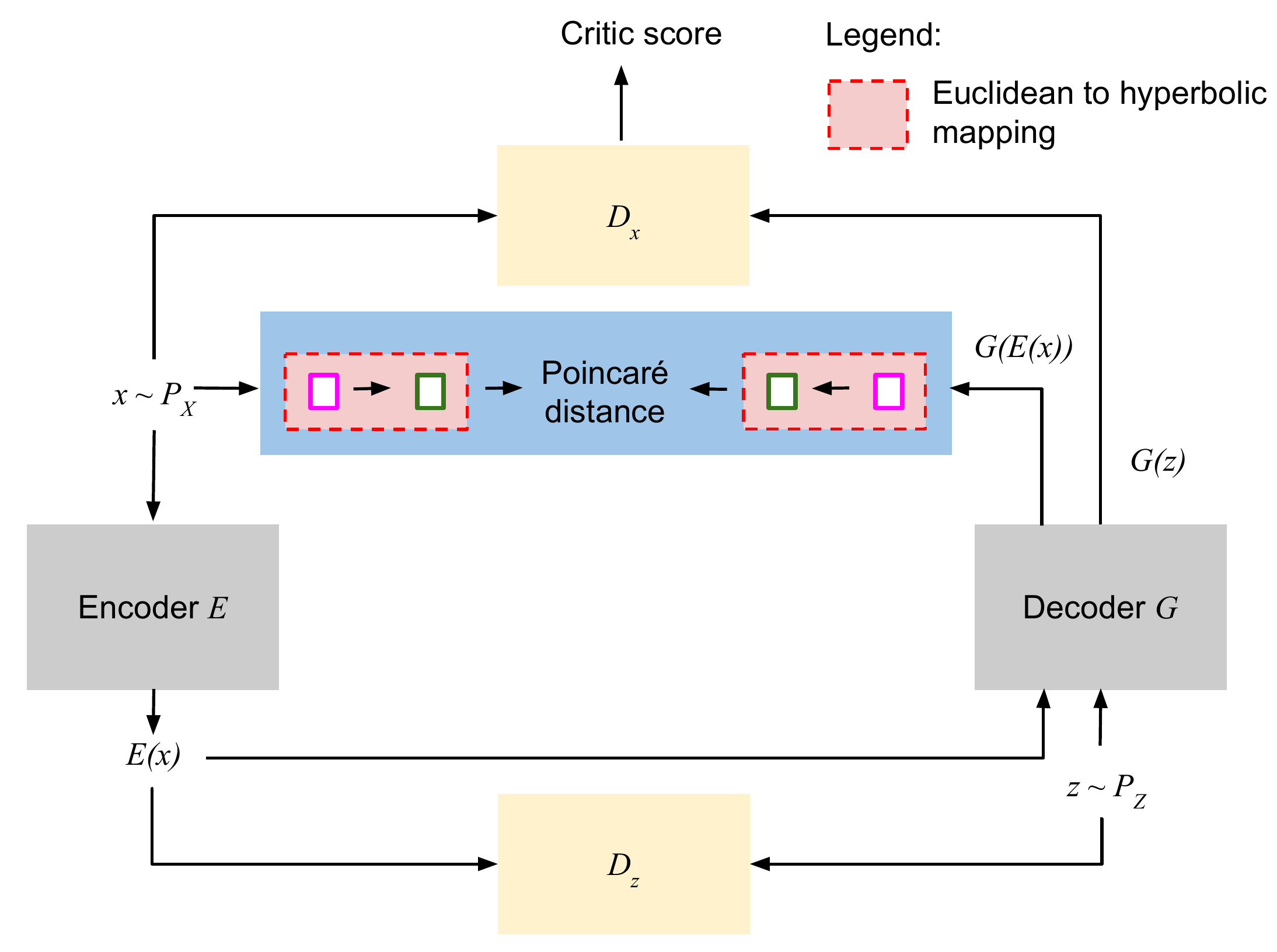}
\end{center}
\vspace*{-5mm}
\caption{Overall architecture of our proposed model HypAD. The model integrates hyperNNs~\cite{NEURIPS2018_dbab2adc} with an LSTM based encoder-decoder trained with two \textit{GAN}-based discriminators, $D_x$ and $D_z$. HypAD maps the input signal $x$ and the output of the decoder ($G(E(x))$) to the Poincar\'e ball model of hyperbolic spaces, shown as dotted red edge box with red background, to get the corresponding hyperbolic embeddings. The two embeddings are then compared using  Poincar\'e distance, as described in Sec.~\ref{sec:hypRe}. Solid magenta and green boxes denote the input and output of the hyperbolic mapping, respectively.}
\vspace*{-4mm}
\label{fig:hypAD}
\end{figure}

%% file: fig/cosine_barplot.tex
\begin{figure*}
\begin{center}
	\includegraphics[width=0.98\textwidth]{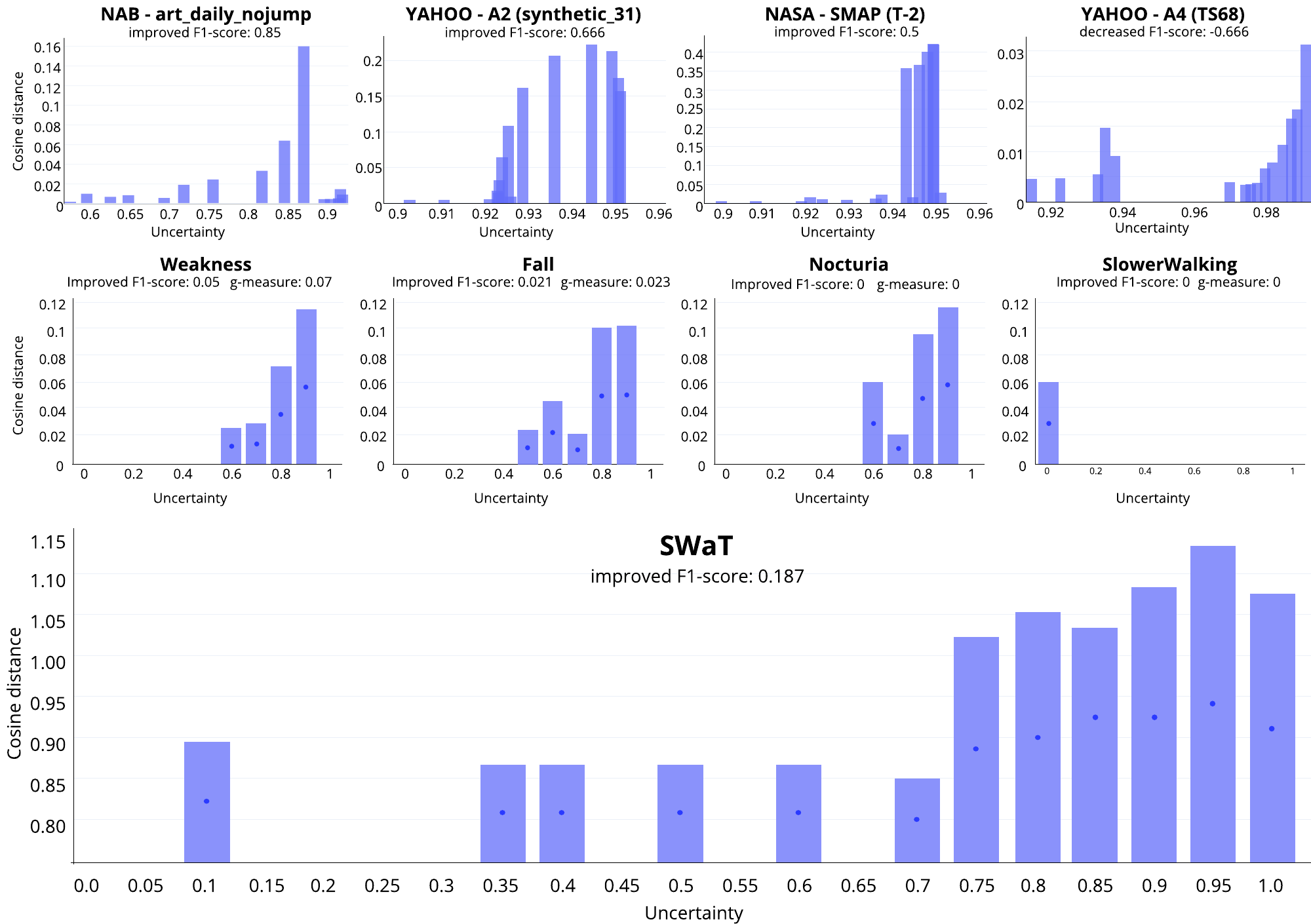}
\end{center}
\vspace*{-5mm}
\caption{Bar plot of the average cosine distance of all the datasets for specific intervals of uncertainty (see Sec. \ref{sec:mot}). The first two rows contain three plots corresponding to the signals that report the best improvement in terms of F1-score (g-measure) and one plot (the last column) that corresponds to the signal with the worst improvement. Notice that even the signals with the worst improvement follow the same increasing trend. Because SWaT contains a single long-term signal, we report only its corresponding bar-plot.} 
\vspace*{-4mm}
\label{fig:cosine_barplot}
\end{figure*}

%% file: tab/datasets.tex
\begin{table*}[ht]
\centering
\resizebox{\textwidth}{!}{%
\begin{tabular}{llccccccccccccccccc}
\multicolumn{2}{l}{\multirow{3}{*}{}} & \multicolumn{11}{c}{Univariate} & \multicolumn{6}{c}{Multivariate} \\ \cline{3-19} 
\multicolumn{2}{l}{} & \multicolumn{2}{c}{NASA} & \multicolumn{4}{c}{Yahoo} & \multicolumn{5}{c}{NAB} & \multicolumn{5}{c}{CASAS} & SWaT\\ \cline{3-19} 
\multicolumn{2}{l}{} & SMAP & MSL & A1 & A2 & A3 & A4 & Art & AdEx & AWS & Traf & Tweets & F & MTC & W & SW & N & \\ \hline
\multicolumn{2}{l}{Num. of signals} & 53 &  27 & 67 & 100 & 100 & 100 & 6 & 5 & 17 & 7 & 10 & 1 & 1 & 1 & 1 & 1 & 1\\
\multicolumn{2}{l}{Num. of anomalies} & 67 & 36 & 178 & 200 & 939 & 835 & 6 & 11 & 30 & 14 & 33 & 2 & 2 & 4 & 2 & 2 & 33\\
 & Point anomalies & 0 & 0 & 68 & 33 & 935 & 833 & 0 & 0 & 0 & 0 & 0 & 0 & 0 & 0 & 0 & 0 & 0\\
 & Collective anomalies & 67 & 36 & 110 & 167 & 4 & 2 & 6 & 11 & 30 & 14 & 33 & 2 & 2 & 4 & 2 & 2 & 33\\
\multicolumn{2}{l}{Num. of anomaly points} & 54,696 & 7,766 & 1,699 & 466 & 943 & 837 & 2,418 & 795 & 6,312 & 1,560 & 15,651 & 99 & 239 & 1,248 & 276 & 1,060 & 10786\\
& Percentage of total & 9.7\% & 5.8\% & 1.8\% & 0.3\% & 0.6\% & 0.5\% & 10\% & 9.9\% & 9.3\% & 9.8\% & 9.8\% & 0.7\% & 1.7\% & 8\% & 2.4\% & 6.3\% & 10.05\%\\

& Num. out of distribution & 18,126 & {642} & 861 & 153 & 21 & {49} & 123 & 15 & 210 & 86 & 520 & - & - & - & - & - & -\\
\multicolumn{2}{l}{Num. of instances} & $\sim563$k & $\sim132$k & $\sim95$k & $\sim142$k & 168k & 168k & $\sim24$k & $\sim8$k & $\sim68$k & $\sim16$k & $\sim159$k & $\sim54$k & $\sim276$k & $\sim39$k & $\sim39$k & $\sim39$k & $\sim189$k\\
\multicolumn{2}{l}{Synthetic?} & No & No & No & Yes & Yes & Yes & Yes & No & No & No & No & No & No & No & No & No & No\\ \hline
\end{tabular}%
}
\caption{Overview of selected univariate and multivariate datasets and their characteristics, grouped by the sources of signals (NASA, Yahoo, NAB, CASAS and SWaT). See Sec.~\ref{sec:dataset} for details.}
\vspace*{-4mm}
\label{tab:datasets}

\end{table*}

%% file: sec/experiments.tex
\section{Results}\label{sec:expres}

We compare HypAD against the best univariate anomaly detector TadGAN~\cite{geiger2020tadgan} on a benchmark of 11-time series and extend the comparison to 2 multivariate sensor datasets: one comprising a water treatment plant and one comprising daily activities in elderly residences.
First, we introduce the benchmarks (Sec.~\ref{sec:dataset}), then we compare against baselines and the state-of-the-art (Sec.~\ref{sec:sota}), finally, we conduct ablative studies on the importance of uncertainty for performance and the reduction of false alarms (Sec.~\ref{sec:abl}).

\subsection{Datasets, metrics and experimental setup}\label{sec:dataset}

Tab.~\ref{tab:datasets} summarizes the main characteristics of the datasets, which we coarsely divide into univariate, main test bed of our baseline TadGAN~\cite{geiger2020tadgan}, and multivariate, to which we extend the comparison. In the table, we report the sources of signals (NASA, Yahoo, NAB, CASAS and SWaT) and the datasets within each source (SMAP, MSL, A1, etc.), cf.\ detailed description later in the section.

In the table, the \textit{number of signals} is the number of time series within the datasets. Note that univariate datasets are composed of multiple signals, while multivariate only comprise single large time series.
The \textit{number of anomalies} counts the instances, within the time series, labeled as anomalous. These are further detailed as \textit{point anomalies}, anomalous values at a specific point in time, or \textit{collective anomalies}, sets of contiguous times that are altogether anomalous. Yahoo is the sole dataset with point anomalies and with \textit{synthetic} sequences A2, A3, and A4.
We also report the \textit{percentage of total} anomalous points and, following \cite{geiger2020tadgan}, for the univariate datasets, the \textit{number of out-of-distribution} points, exceeding the means by more than $4\sigma$.

\input{tab/performances_univariate}
\input{tab/performances_casas}

\noindent\textbf{Univariate datasets.} \textit{NASA} includes two spacecraft telemetry datasets
based on the Mars Science Laboratory (MSL) and the Soil Moisture Active Passive (SMAP) signals. The former consists of scientific and housekeeping engineering data taken from the Rover Environmental Monitoring Station aboard the Mars Science Laboratory. The latter includes measurements of soil moisture and freeze/thaw state from space for all non-liquid water surfaces globally within the top layer of the Earth.
The \textit{Yahoo} datasets are based on real production traffic to Yahoo computing systems. Additionally, we consider three synthetic datasets coming from the same source.
The dataset tests the detection accuracy of various anomaly types including outliers and change points. The synthetic dataset consists of time-series with varying trends, noise and seasonality. The real dataset consists of time-series representing the metrics of various Yahoo services.
\textit{Numenta Anomaly Benchmark (NAB)} is a well-established collection of univariate time-series from real-world application domains.
To be consistent with \cite{geiger2020tadgan}, we analyse Art, AdEx, AWS, Traf, and Tweets from the original collection.

\noindent\textbf{Multivariate datasets.} We consider for analysis the SWaT~\cite{10.1007/978-3-319-71368-7_8,Mathur2016SWaTAW} and CASAS~\cite{cook2012casas,dahmen2021indirectly} datasets.
\textit{SWaT} is collected from a cyber-physical system testbed that is a scaled-down replica of an industrial water treatment plant. The data was collected every second for a total of 11 days, where for the first few days the system was operated normally, while for the remaining days, certain cyber-physical attacks were launched. Following~\cite{10.1145/3447548.3467137}, we sample sensor data every 5 seconds.
\textit{CASAS} is a collection of two weeks of sensor data from retirement homes. Each sensor reading has a label attached to it, according to the activity of elderly people recognised by human annotators. The 5-time series are a collection of activities, organised by the pre-established medical conditions: Falling, More Time in Chair, Weakness, Slower Walking, and Nocturia for nightly time visits.
Although sensor readings give fine-grained information, we are interested in creating daily patient profiles. Hence, we collapse them into a single engendered activity with a start and end time
for each consecutive sensor signal. We then create a time matrix structure $\mathcal{M}^{|d|,1440}$ where $|d|$ is the number of days the patient is monitored and 1440 represent the total number of minutes in a day. $\mathcal{M}[i,j]$ represents the label performed in minute $j$ of day $i$. Lastly, we densify each value $\mathcal{M}[i,j]$ with contextual and duration information corresponding to the label therein.

Finally, we create train and test splits of this data, safeguarding the sequentiality of observations, and gathering the few anomalies into the test set for evaluation only. We name this Unsupervised-CASAS, dubbed U-CASAS, differing from the original CASAS \cite{cook2012casas,dahmen2021indirectly} since the latter interleaves anomalous sensor readings with normal instances, thus breaking the sequentiality of anomalies.
For each anomalous day encountered in the test set, we pad two days prior to it and two after. If the padding overlaps with another anomalous day, then we concatenate\footnote{The concatenation procedure merges common sequences. If the sequence contains more than one anomalous day within the two-day padding window, the padded sequence gets collapsed into a single one.} them and perform the padding procedure again. We delete the days assigned to the test set from the overall dataset and assign the rest as training. Moreover, since we employ a time-related strategy, we create time windows of 30 actions to detect anomalies.

\noindent\textbf{Metrics.}
Being all the enlisted datasets highly unbalanced, accuracy is misleading. Therefore, as done in \cite{suris2021hyperfuture}, we use the F1 score to account for this challenge. Notice that we do not use the cumulative F1 score as proposed in \cite{garg2021evaluation} to evaluate the performances because not all datasets contain anomalous events; rather they contain anomalous data points. Based on \cite{hundman2018detecting}, for the univariate and the CASAS datasets, we penalize high false positive rates and encourage the detection of true positives in a timely fashion. Since anomalies are rare events and come in collective sequences in real-world applications, we proceed as follows:
\begin{enumerate}[noitemsep,topsep=0pt] 
\item We record a true positive (TP) if any predicted window overlaps a true anomalous window.
\item We record a false negative (FN) if a true anomalous window does not overlap any predicted window.
\item We record a false positive (FP) if a predicted window does not overlap any true anomalous region.
\end{enumerate}
For U-CASAS, we also measure the g-measure which is the geometric mean of the product of recall (R) and precision (P), being a robust metric when classes are imbalanced \cite{dahmen2021indirectly}.

\noindent\textbf{Baselines.}
 We include the following strategies as our baselines in this paper:
\begin{itemize}[noitemsep,topsep=0pt]
    \item \textit{AE} \cite{baldi2012autoencoders} - We use a six-layer dense autoencoder.
    \item \textit{ConvAE} \cite{maggipinto2018convolutional} - We have three layers of convolutional encoding interleaved with max pooling. The decoder has a specular composition as the encoder where the de-convolution is aided by two-dimensional up-sampling layers.
    \item \textit{LstmAE} \cite{sagheer2019unsupervised} - We use a deep-stacked LSTM autoencoder with four layers. The first LSTM hidden and output vectors get passed to the second LSTM layer. The latent representation of the encoder gets then reconstructed in reverse order from the decoder.
    \item \textit{TadGAN} \cite{geiger2020tadgan} - We use a one-layer bidirectional LSTM for the generator $E$, and a two-layer bidirectional LSTM for $G$. For the critic $D_z$ we use a fully connected layer, and two dense layers for $D_x$.
\end{itemize}

\noindent\textbf{Implementation details.} For the first three baselines, we set the number of epochs to $30$, the batch size to $32$, and the learning rate to $10^{-3}$. For TadGAN\footnote{Pytorch implementation available at \url{https://github.com/arunppsg/TadGAN}}, we set the epochs to $30$, the batch size to $64$, the window width to $100$, the learning rate to $5\times10^{-4}$, and the iteration for the critic to 5. We use Adam as the optimizer to train all the baselines. For our proposed method HypAD, we took inspiration from the PyTorch implementation of TadGAN. We leave the architecture of TadGAN unvaried, and incorporate the hyperbolic transformation as in \cite{suris2021hyperfuture}. The hyperparameters are the same as in the original paper, using Riemannian Adam as optimizer.

\subsection{Comparison to the state of the art} \label{sec:sota}

HypAD defines a new state-of-the-art performance for univariate anomaly detection by having the highest average F1-scores of 0.631.
In Tab.~\ref{tab:performances_univariate}, HypAD outperforms the current best technique, TadGAN, by 5.17\%, as well as all baselines by a large margin.
In the Table, the column \textit{F1} reports mean $\mu$ and standard deviation $\sigma$ over all datasets.  Looking at $\sigma=0.075$, HypAD also appears as the most consistent performer.
Considering the F1-score, the largest performance gain of HypAD Vs TadGAN are on NAB and NASA datasets, while it is outperformed more largely on the A2, A3 and A4 Yahoo datasets, the synthetic ones.

In Tab.~\ref{tab:performances_casas}, we extend the evaluation of HypAD to the multivariate U-CASAS dataset. We cannot include Isudra~\cite{dahmen2021indirectly} because the underlying architecture uses a small amount of labels to select parameters for the execution. Additionally, Isudra is trained in a supervised fashion differing from all the other methods reported here. For completeness, we report the g-measure~\cite{dahmen2021indirectly} and its average across all datasets. 
As shown in the table, HypAD surpasses TadGAN by 32.51\% in terms of the average F1-score.

Finally, in Tab.~\ref{tab:performances_swat_fg}, we extend the comparison to the multivariate SWaT dataset. Here, following~\cite{10.1145/3447548.3467137}, we report the precision and recall in addition to the F1-score. HypAD outperforms the baseline TadGAN (0.753 Vs.\ 0.722), but both techniques are behind the current SoA on multivariate anomaly detection, NSIBF~\cite{10.1145/3447548.3467137}.
HypAD achieves its highest F1 performance at the highest precision (0.996) among all other methods. This confirms that HypAD detects anomalies which it is certain about, i.e.\ when it \emph{understands} the time series and it is certain about what to expect, but cannot reconstruct the input signal due to an anomaly.

\input{tab/swat_permormances}

\input{fig/ablation1.tex}

\subsection{Ablation Studies}\label{sec:abl}

In Tab.~\ref{tab:ablation}, we analyze the importance of the hyperbolic embedding and the use of uncertainty for anomaly detection on the univariate datasets, as well as the multivariate U-CASAS and SWaT.
The first row shows the performance of the model in Euclidean space (average F1-score across datasets). This corresponds to TadGAN in Tab.~\ref{tab:performances_univariate},~\ref{tab:performances_casas} and~\ref{tab:performances_swat_fg}. In the second row, we report the performance for TadGAN with the inclusion of uncertainty via 100 Monte Carlo dropouts. In the third row, we report performance for the hyperbolic TadGAN without including uncertainty (cf.~Sec.~\ref{sec:hypRe}). This improves marginally over the univariate (0.604 Vs.\ 0.600) and, more importantly, over the U-CASAS datasets (0.397 Vs.\ 0.323), but it decreases performance in SWaT (0.566 Vs.\ 0.722), which we further analyze in the following subsection. 
The complete proposed HypAD model, in the fourth row, improves consistently on both ablative variants. HypAD yields a 4.5\% gain on the univariate datasets, 7.8\% on U-CASAS, and 33\% on SWaT. We infer that hyperbolic mapping and its associated uncertainty are fundamental for ameliorating anomaly detection over the Euclidean counterparts. 

 \input{tab/ablation}

\noindent \textbf{Qualitative Ablation on SWaT.} In Fig.~\ref{fig:ablation1}, we demonstrate qualitative ablation on the SWaT dataset, which consists of a single long signal. The three plots correspond to the three ablative variants of Tab.~\ref{tab:ablation}. In all plots, the blue and green points represent the predicted anomaly scores and the ground-truth anomalies, respectively. The red line denotes the anomaly detection threshold i.e.\ blue points above the red line are the predicted anomalies. FP are blue points above the red line threshold outside the green (ground-truth) anomalous regions. As shown in the figure, the Euclidean model (first plot) yields many FP; the hyperbolic model w/o uncertainty (second plot) reduces the number of FP substantially, but it also misses anomalies, esp.\ in the middle signal part. This explains the drop in the F-1 score for the SWaT dataset (cf.\ Table~\ref{tab:ablation}). Integrating hyperbolic uncertainty (proposed HypAD, third plot) recovers the detection of these anomalies. It increases TP but maintains the number of FP low, thus yielding the best F1 score of 0.753.

%% file: tab/performances_univariate.tex
\begin{table*}[t]
\centering
\resizebox{0.95\textwidth}{!}{%
\begin{tabular}{@{}llcc|cccc|ccccc|c@{}}
\multicolumn{2}{l}{\multirow{2}{*}{}} & \multicolumn{2}{c}{NASA} & \multicolumn{4}{c}{YAHOO} & \multicolumn{5}{c}{NAB} &  \\ \cline{3-14} 
\multicolumn{2}{l}{} & MSL & SMAP & A1 & A2 & A3 & A4 & Art & AdEx & AWS & Traf & Tweets & F1 ($\mu\pm\sigma$) \\ \cline{1-14} 
\multicolumn{2}{l}{AE} & 0.199 & 0.270 & 0.283 & 0.008 & 0.100 & 0.073 & 0.283 & 0.100 & 0.239 & 0.088 & 0.296 & 0.176 $\pm$ 0.099 \\
\multicolumn{2}{l}{LstmAE} & 0.317 & 0.318 & 0.310 & 0.023 & 0.097 & 0.089 & 0.261 & 0.130 & 0.223 & 0.136 & 0.299 & 0.200 $\pm$ 0.103 \\
\multicolumn{2}{l}{ConvAE} & 0.300 & 0.292 & 0.301 & 0.000 & 0.103 & 0.073 & 0.289 & 0.129 & 0.254 & 0.082 & 0.301 & 0.212 $\pm$ 0.096 \\
\multicolumn{2}{l}{TadGAN}                    & 0.500 & 0.580 & \textbf{0.620} & \textbf{0.865} & \textbf{0.750} & \textbf{0.576} & 0.420 & 0.550 & \textbf{0.670} & 0.480 & 0.590 & 0.600 $\pm$ 0.115 \\
\multicolumn{2}{l}{HypAD (proposed) } & \textbf{0.565} & \textbf{0.643} & {0.610} & 0.670 & 0.670 & 0.470 & \textbf{0.777} & \textbf{0.663} & 0.630 & \textbf{0.570} & \textbf{0.670} & \textbf{0.631 $\pm$ 0.075} \\  \cline{1-14} 
\end{tabular}%
}
\caption{Results on the univariate datasets from NASA, YAHOO and NAB, measured in terms of F1-score.
}
\vspace*{-2mm}
\label{tab:performances_univariate}

\end{table*}

%% file: tab/performances_casas.tex
\begin{table*}[t]
\centering

\resizebox{\textwidth}{!}{%
\begin{tabular}{lcc|cc|cc|cc|cc|cc}
\multirow{2}{*}{} & \multicolumn{2}{c}{Fall} & \multicolumn{2}{c}{Weakness} & \multicolumn{2}{c}{Nocturia} & \multicolumn{2}{c}{Slower-walking} & \multicolumn{2}{c}{More time in chair} & \multicolumn{2}{c}{} \\ \cline{2-13} 
 & G & F1 & G & F1 & G & F1 & G & F1 & G & F1 & G ($\mu \pm \sigma$) & F1 ($\mu \pm \sigma$)\\ \hline
LstmAE & 0.085 & 0.014 & 0.182 & 0.108 & 0.000 & 0.000 & 0.158 & 0.049 & 0.133 & 0.035 & 0.112 $\pm$ 0.064 & 0.041 $\pm$ 0.037 \\

AE & 0.139 & 0.127 & 0.033 & 0.027 & 0.116 & 0.103 & 0.000 & 0.000 & 0.158 & 0.049 & 0.089 $\pm$ 0.062 & 0.061 $\pm$ 0.047 \\

ConvAE & 0.086 & 0.014 & 0.284 & 0.150 & 0.251 & 0.119 & 0.158 & 0.048 & 0.134 & 0.035 & 0.183 $\pm$ 0.074 & 0.073 $\pm$ 0.052 \\
TadGAN & 0.222 & 0.267 & 0.570 & 0.555 & 0.000 & 0.000 & \textbf{0.630} & \textbf{0.570} & 0.267 & 0.222 & 0.338 $\pm$ 0.233 & 0.323 $\pm$ 0.216 \\
HypAD (proposed) & \textbf{0.447} & \textbf{0.333} & \textbf{0.660} & \textbf{0.610} & \textbf{0.447} & \textbf{0.333} & 0.470 & 0.364 & \textbf{0.577} & \textbf{0.500} & \textbf{0.520 $\pm$ 0.095} & \textbf{0.428 $\pm$ 0.123}\\ \cline{1-13} 
\end{tabular}%
}
\caption{Results for the U-CASAS multivariate datasets, measured in terms of F1-score and G-measure. }
\vspace*{-6mm}
\label{tab:performances_casas}

\end{table*}

%% file: tab/swat_permormances.tex

\begin{table}[t]
\centering
\resizebox{\linewidth}{!}{%
\begin{tabular}{lccc}
\textbf{Model}                    & \textbf{Precision} & \textbf{Recall} & \textbf{F1} \\ 
\hline
EncDec-AD~\cite{Malhotra2016LSTMbasedEF} {[}ICML-WorkShop16{]} & 0.945        & 0.620        & 0.748       \\
DAGMM~\cite{zong2018deep}{[}ICLR18{]}              & 0.946        & 0.747        & 0.835       \\
OmniAnomaly~\cite{10.1145/3292500.3330672} {[}KDD19{]}         & 0.979        & 0.757        & 0.854       \\
USAD~\cite{10.1145/3394486.3403392} {[}KDD20{]}                & 0.987        & 0.740        & 0.846       \\
TadGAN {[}BigData20{]}          & 0.937        & 0.587        & 0.722       \\
NSIBF~\cite{10.1145/3447548.3467137} {[}KDD21{]}              & 0.982        & \textbf{0.863}        & \textbf{0.919}      \\
HypAD (Ours)                  & \textbf{0.996}        & 0.605        & 0.753    \\
\hline
\end{tabular}%
}
\caption{Results on the SWAT dataset, in terms of F1-score, including the corresponding precision and recall.
}
\label{tab:performances_swat_fg}

\end{table}

%% file: fig/ablation1.tex
\begin{figure} 
\vspace*{-2mm}
\begin{center}
	\includegraphics[width=0.49\textwidth]{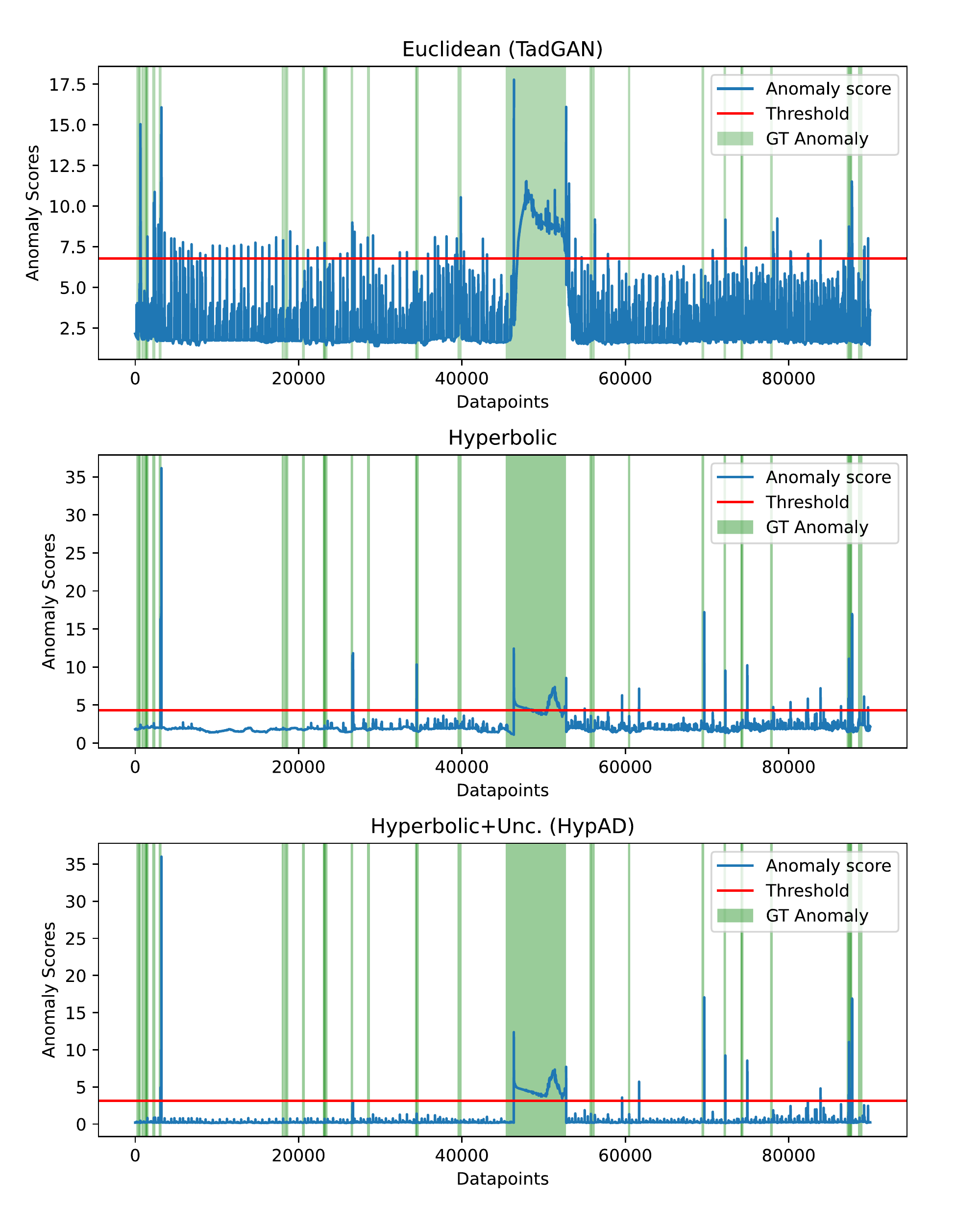}
\end{center}
\vspace*{-5mm}
\caption{Qualitative Ablation on SWaT: Anomaly scores (blue points) that lie above the anomaly detection threshold (red line) but do not coincide with the ground-truth (green points) are FP. Euclidean model in plot 1 has many FP. The corresponding hyperbolic model in plot 2 is more precise but loses some TP, especially in the middle region. Integration of uncertainty helps to recover these points, increasing the TP and overall F-1 score (Sec~\ref{sec:abl}).}
\label{fig:ablation1}
\end{figure}

%% file: tab/ablation.tex
\begin{table}[t]
\centering
\resizebox{\linewidth}{!}{%
\begin{tabular}{lccc}
 & Univariate & U-CASAS & SWaT\\ \hline
{Euclidean (TadGAN)} &  0.600 & 0.323 & 0.722 \\

{Euclidean (TadGAN) + MC dropout} &  0.618 & 0.339 & 0.734 \\

{Hyperbolic}  & 0.604 & 0.397 & 0.566 \\

{Hyperbolic + Uncertainty (HypAD)}&  \textbf{0.631} & \textbf{0.428} & \textbf{0.753}\\ \hline
\end{tabular}%
}
\caption{Ablative evaluation on the importance of the hyperbolic mapping and the integration of uncertainty. Average F1-scores are reported for Univariate, U-CASAS and SWaT datasets.
}
\vspace*{-5mm}
\label{tab:ablation}

\end{table}




%% file: sec/conclusion.tex
\section{Conclusions}
We have proposed a novel model for anomaly detection based on hyperbolic uncertainty, HypAD. 
The proposed hyperbolic uncertainty allows HypAD to self-adjust its output, encouraging the model to either predict a correct reconstruction or a less certain wrong one. This benefits anomaly detection in two ways: it provides better reconstructions of the signal (the deviation from those being anomalous) and it yields a measure of certainty.
This is a novel viewpoint on anomaly detection: detectable anomaly instances are those which are certain but wrongly predicted.